# Simulation-Based Sensitivity Analysis in Optimal Treatment Regimes and Causal Decomposition with Individualized Interventions

Soojin Park[1]

Suyeon Kang[2]

Chioun Lee[3]

[1]School of Education, University of California, Riverside

[2]Department of Statistics and Data Science, University of Central Florida

[3]Department of Sociology, University of California, Riverside

## Author Note

Research reported in this publication was supported by the National Science Foundation (NSF, 2243119) and the National Institute on Aging of the National Institutes of Health (NIH, R24AG077433). Correspondence concerning this article should be addressed to Soojin Park, Graduate School of Education, The University of California-Riverside, Riverside, CA92521, USA. Email: soojin.park@ucr.edu

## Abstract

Causal decomposition analysis aims to assess the effect of modifying risk factors on reducing social disparities in outcomes. Recently, this analysis has incorporated individual characteristics when modifying risk factors by utilizing optimal treatment regimes (OTRs). Since the newly defined individualized effects rely on the no omitted confounding assumption, developing sensitivity analyses to account for potential omitted confounding is essential. Moreover, OTRs and individualized effects are primarily based on binary risk factors, and no formal approach currently exists to benchmark the strength of omitted confounding using observed covariates for binary risk factors. To address this gap, we extend a simulation-based sensitivity analysis that simulates unmeasured confounders, addressing two sources of bias emerging from deriving OTRs and estimating individualized effects. Additionally, we propose a formal bounding strategy that benchmarks the strength of omitted confounding for binary risk factors. Using the High School Longitudinal Study 2009 (HSLS:09), we demonstrate this sensitivity analysis and benchmarking method.

Simulation-Based Sensitivity Analysis in Optimal Treatment Regimes and Causal Decomposition with Individualized Interventions

## 1. Introduction

Substantial disparities in educational, economic, and health outcomes persist across social groups in the United States, including those based on race/ethnicity, class, and gender. Traditional decomposition approaches, such as difference-in-coefficients analysis or Oaxaca-Blinder decomposition, have been employed to identify risk factors, such as educational attainment, that explain these disparities. However, these approaches have limitations: they do not define clear causal estimands and fail to specify the assumptions required for causal interpretation (e.g., no omitted confounding). Causal decomposition analysis (VanderWeele & Robinson, 2014; Jackson & VanderWeele, 2018) has addressed these issues by defining causal estimands, such as disparity reduction and disparity remaining, and by clarifying assumptions within a counterfactual framework.

Disparity reduction and disparity remaining are defined as the extent to which social disparities (e.g., racial disparities in math scores) would be reduced or remain after hypothetically intervening to set risk factors (e.g., taking Algebra I by 9th grade) to a pre-specified value (Lundberg, 2020) or equalize the distribution of risk factors across groups (Jackson & VanderWeele, 2018). However, these hypothetical interventions often overlook individual characteristics (e.g., prior achievement or interest in math), which limits their real-world applicability. To address this limitation, Park, Lee, and Quintana (2024a) incorporated individual characteristics when intervening on risk factors by using Optimal Treatment Regimes (OTRs). OTRs are decision rules that guide the assignment of treatments based on individual characteristics, with the goal of maximizing the expected outcome through leveraging heterogeneous effects (Murphy, 2003). Utilizing this concept, Park et al. (2024a) defined the individualized effects of disparity reduction and disparity remaining, aiming to investigate whether the intervention that maximizes the desired outcome also reduces disparities in the outcome between groups. The newly developed causal decomposition framework can serve as a critical tool for policymakers and educators to design interventions that balance excellence (effectively enhancing the desired outcome) and equity (reducing disparities between groups).

These newly developed individualized effects defined by Park et al. (2024a), similar to other causal estimands, rely on unverifiable assumptions, such as no omitted confounding. Therefore, it is essential to develop a sensitivity analysis to assess the robustness of individualized effects against potential violations of the no omitted confounding assumption. While the bias originates from the presence of potential omitted variables, these variables may impact the estimation of both OTRs and disparity reduction/remaining. A difficulty in developing such a sensitivity analysis is uncertainty about whether these impacts are independent of one another. Although several studies have developed sensitivity analyses in the context of causal decomposition analysis (e.g., Park, Qin, & Lee, 2022; Park, Kang, Lee, & Ma, 2023; Rubinstein, Branson, & Kennedy, 2023), this specific challenge cannot be easily addressed with existing formula-based sensitivity analysis. Additionally, while the instrumental variables (IVs) approach has often been employed to address omitted confounding in the OTR literature, (e.g., Cui & Tchetgen Tchetgen, 2021; Qiu et al., 2021), identifying suitable IVs in observational data remains a significant challenge, limiting their applicability.

By extending the simulation-based sensitivity analysis introduced by Carnegie, Harada, and Hill (2016), we aim to address the challenge of assessing the sensitivity of OTRs and individualized effects. The proposed approach simulates an omitted confounder given observed variables, allowing us to address both with and without interactions between the omitted variable and the risk factor. However, due to inferior performance observed in the simulation study, we focus on the case without interactions. Further discussion is provided in Section 6.

Most importantly, we enhance the sensitivity analysis by providing a method to formally benchmark the strength of omitted confounding. Sensitivity analyses typically evaluate the amount of omitted confounding required to alter study conclusions and whether such amounts of confounding are plausible within the research context. However, determining the plausibility of the amount can be difficult due to the lack of precise knowledge about it. To tackle this issue, Cinelli and Hazlett (2020) developed a novel bounding strategy to benchmark the strength of omitted confounding against observed covariates using $R^2$. However, this approach is not suitable for simulation-based sensitivity analysis, especially when the risk factor is binary and the outcome is continuous. Therefore, we propose a strategy that benchmarks the strength of unmeasured confounding against observed covariates using original data scales, making it suitable for simulation-based sensitivity analysis even with binary risk factors.

To summarize, this study has two primary objectives: 1) to extend simulation-based sensitivity analysis to causal decomposition with individualized interventions, and 2) to enhance the sensitivity analysis by developing a method to benchmark omitted confounding against observed covariates applicable to simulation-based sensitivity analysis with binary risk factors. The structure of the remaining sections is as follows: Section 2 introduces a running example and Section 3 provides a review of causal decomposition with individualized interventions. Section 4 introduces a simulation-based sensitivity analysis that is applicable to individualized interventions and demonstrates its performances through simulation studies. Section 5 proposes a method for benchmarking the strength of omitted confounding against observed covariates. Finally, we conclude with a discussion of the main contributions of the paper. Open-source software for R (`causal.decomp`) that implements the proposed method is available from https://cran.r-project.org/web/packages/causal.decomp/index.html.

## 2. Running Example

Our study is motivated by the following question: “How much of the Black-White disparity in math achievement would be reduced or remain if we intervened in students’ enrollment in Algebra I in 9th grade?" This example is drawn from the previous paper by Park, Lee, and Quintana (2024b). To explore this question, we used data from the High School Longitudinal Study 2009 (HSLS:09).

Although most U.S. students take Algebra I in the ninth grade, there are notable racial and ethnic differences in the timing of enrollment. Non-Hispanic White and Asian students are more likely to enroll before ninth grade, whereas Hispanic and Black students are more likely to take the course in ninth grade or later (National Center for Education Statistics, 2019). Earlier completion of Algebra I can help students advance to higher-level math courses, such as Algebra II, Geometry, and Pre-calculus, earlier in high school (Cohen & Hill, 2000). For example, "Algebra for All" policies aim to increase access to advanced mathematics by mandating Algebra I for all students by a specific grade level (Silver, 1995). However, universal eighth-grade Algebra I enrollment has sparked debates, as some argue that it can help to close racial and socioeconomic achievement gaps (Stein, Kaufman, Sherman, & Hillen, 2011) while it also presents challenges related to student readiness, equity, and long-term outcomes (Loveless, 2008; Chazan, Sela, & Herbst, 2016; Clotfelter, Ladd, & Vigdor, 2015).

Instead of mandating universal eighth-grade Algebra I enrollment, a more flexible and student-centered approach has been recommended in recent years. For most students, taking Algebra I in ninth grade is appropriate, as it aligns with traditional college-preparatory math sequences. However, school districts could offer accelerated tracks for students who demonstrate readiness before ninth grade, ensuring they progress at a pace suited to their abilities (Domina, McEachin, Penner, & Wimberly, 2015). As an example, our study considers Algebra I enrollment by ninth grade (i.e., taking Algebra I by ninth grade based on a student's readiness) as a key factor in reducing educational inequalities. Our intervention strategy focuses on students who have not taken Algebra I before 9th grade while allowing those who took it earlier to remain unchanged. By applying causal decomposition analysis, we assess the impact of this intervention on achievement disparities, offering insights into the impact of such enrollment policies.

***Directed acyclic graph (DAG).*** We consider social groups represented by two categories: Blacks ($R = 1$) and Whites ($R = 0$). Our outcome of interest is the math score in 11th grade ($Y$). The risk factor is whether students took Algebra I ($M$) in 9th grade. As taking Algebra I in 9th grade is not a random process, its relationship with math score is confounded by multiple factors. Therefore, based on literature (Byun, Irvin, & Bell, 2015; Kelly, 2009; Long, Conger, & Iatarola, 2012; Riegle-Crumb & Grodsky, 2010), we identified confounders, such as gender ($C$), childhood socioeconomic status (SES, $X_1$), and student, parental, and school characteristics that are related to math readiness ($X_2$), such as students' grades, math interest, parental expectations and aspirations regarding their children. We also recognize the presence of unmeasured variables ($U$) that may confound the $M - Y$ relationship.

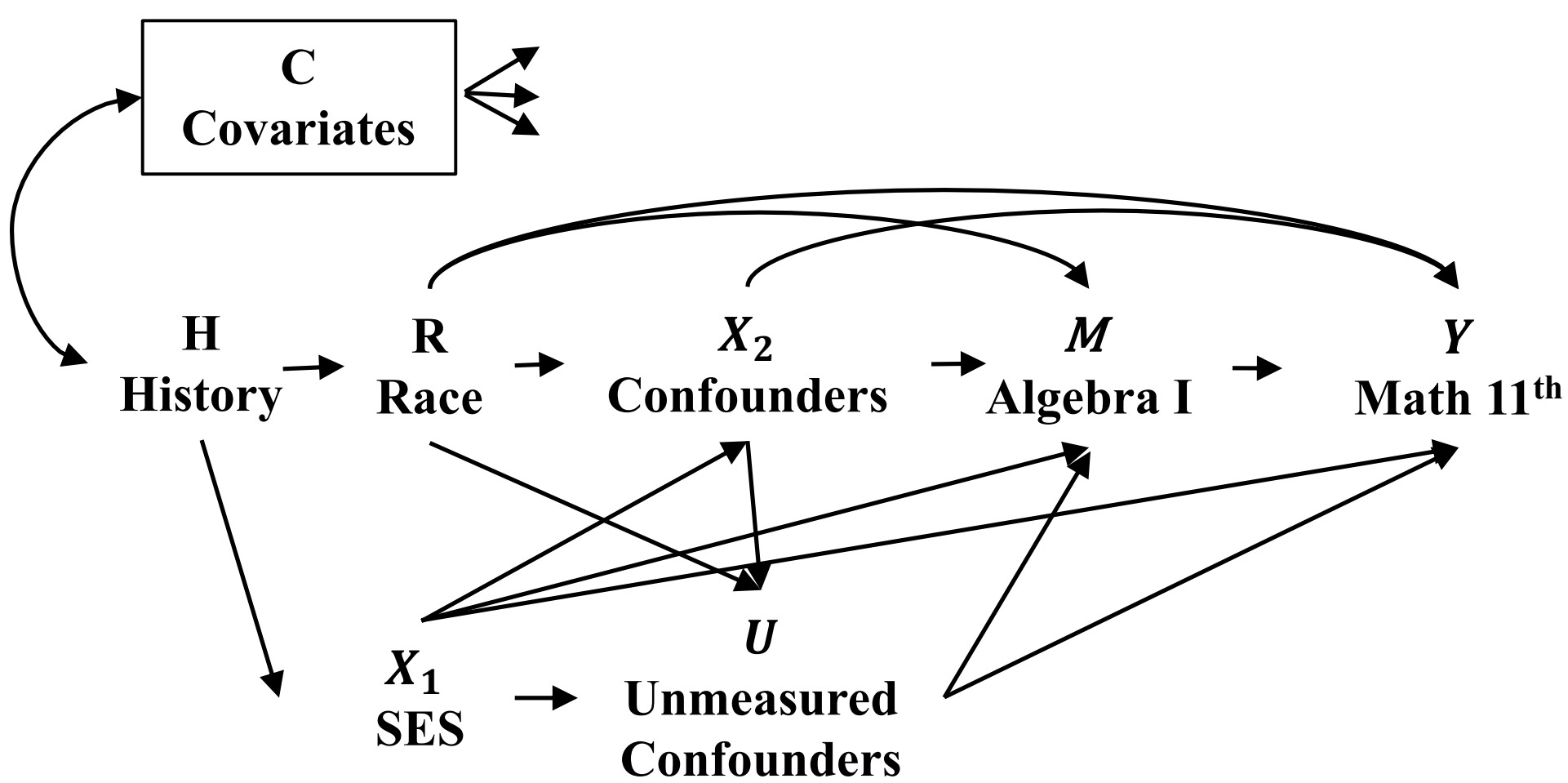

*Figure 1*. DAG showing the pathways to the racial disparity in math achievement in 11th grade

Note. 1) Baseline covariates ($C$) include gender, and placing a box around $C_i$ indicates conditioning on this variable. 2) The three arrows emanating from $C$ indicate that they are confounders of all bivariate relations, as visualized by the box around $C$. 3) Childhood SES ($X_1$) includes the individual's birth region, and father's and mother's years of education. 4) Confounders ($X_2$) include student, parental, and school characteristics that are related to math readines.

To illustrate these relationships, we draw a DAG as shown in Figure 1. We assume that the Black-White math disparity arises through these three paths : (P1) the path from race to math score through taking Algebra I by 9th grade (e.g., $R \to (X_2, U) \to M \to Y$), (P2) the backdoor path from race to math score through historical processes (e.g., $R \to H \to (X_1, X_2, U) \to M \to Y$), and (P3) the remaining path from race to math score not through taking Algebra I by 9th grade. The path (P1) represents when Black students are more likely to be assigned to a less rigorous curriculum track as a result of discrimination and unequal access to educational resources and opportunities (Kelly, 2009), thereby contributing to Black-White math disparities. The path (P2) depicts the situation where Black students are more likely to be born into families with lower SES as a result of historical processes that involve racism, such as slavery and segregation (Kaufman, 2008; Jackson & VanderWeele, 2018). In all paths, the effect of taking Algebra I by 9th grade on math achievement in 11th grade will be biased if unmeasured confounders $U$ exist.

***Sample and Measures.*** We restricted our sample to students who had not taken Algebra I before 9th grade. This restriction ensures that intermediate confounders ($X_2$) measured in 9th grade were assessed before the target factor ($M$). However, this limitation may lead to an underestimation of the initial disparity. After applying this limitation, the sample size was reduced from 14,530 to 11,050, comprising 8,920 White students and 2,130 Black students. Within the restricted data, we used predictive mean matching (Rubin, 1986) to impute missing data, and our estimates are based on a single imputed dataset. Math achievement in 11th grade ($Y$) was measured using the item response theory theta scores. Algebra I ($M$) was defined as students taking Algebra I in 9th grade. For baseline covariates ($C$), we included gender.

Previous literature suggests that performance in earlier courses is the main factor influencing students' course enrollment decisions (Kalogrides & Loeb, 2013). Therefore, we selected students' grades in their most advanced 8th grade math course as an intermediate confounder. Students' enrollment decisions can be influences by parental factors (Byun et al., 2015). Therefore, we also included childhood SES, parents' expectations, and their aspirations regarding their children as intermediate confounders. At the class level, we controlled for the academic disposition of the closest friend. Finally, at the school level, we controlled for school's region, school climate, and the percentage of students in math course that are unprepared, the science and math course requirement, whether the school offers STEM extracurricular activities. We denote intermediate confounders as $X = (X_1, X_2)$.

Although we controlled for an extensive set of confounders, there may still be omitted variables that could confound the relationship between taking Algebra I in 9th grade and math achievement in 11th grade. For example, For example, childhood environment and health conditions can influence cognitive development and academic performance, potentially affecting students' course enrollment decisions and their subsequent math achievement in high school. Therefore, omitted confounders may include variables related to childhood factors, which we denote as $U$ in Figure 1. Given this limitation, conducting a sensitivity analysis is essential to account for potential violations of the no omitted confounding assumption.

## 3. Reducing Disparities with Individualized Interventions: Review

In this section, we review OTRs and the effects of individualized interventions at aiming to reduce disparities. To identify OTRs and evaluate the effects of these interventions on reducing disparities using observational data, we assume the following:

- **A1. Conditional independence of $M$**: This assumption is denoted by $Y(m) \perp M|R = r, X = x, C = c$ for all $x \in \mathcal{X}, c \in \mathcal{C}$ and $r, m = 0, 1$ where $Y(m)$ is a potential outcome under $M = m$. It states that there are no omitted confounders influencing the risk factor-outcome relationship, given the group status, intermediate confounders, and baseline covariates.

- **A2. Positivity**: This assumption is represented as $0 < P(m|r, x, c) < 1$ for all $x \in \mathcal{X}, c \in \mathcal{C}$ and $r, m = 0, 1$. It asserts that individuals of every group have a non-zero probability of experiencing any level of the risk factor (Algebra I) given the group status, intermediate confounders, and baseline covariates.

- **A3. Consistency**: This assumption is represented as $E[Y(m)|r, x, m, c] = E[Y|r, x, m, c]$ for all $x \in \mathcal{X}, c \in \mathcal{C}$ and $r, m = 0, 1$. It indicates that the observed outcome (math score in 11th grade) of an individual with a certain level of the risk factor (Algebra I) is identical to the potential outcome after intervening to set the risk factor to that level.

These three assumptions are quite strong and the plausibility of the assumptions depends on the research context. In our example, consistency could potentially be violated, as the intervention all students to take Algebra I by 9th grade might alter the effect of Algebra I on math achievement. For example, some schools may dilute the curriculum or adjust the level of their instructional rigor (Rosenbaum, 1999), which could differently impact outcomes for students across various schools. Additionally, positivity could be nearly violated if very few students with certain combinations of covariates take Algebra I by 9th grade. However, the impact of violating positivity can be mitigated through methods such as truncation or trimming (Robins, Hernan, & Brumback, 2000). For the purpose of this study, we assume that these assumptions hold. Our primary concern is the conditional independence of risk factor $M$, which is empirically unverifiable and may not hold even after accounting for existing covariates. In Sections 4 and 5, we introduce a sensitivity analysis that enhance the robustness of findings against potential violations of this conditional independence assumption.

### 3.1. Optimal Treatment Regimes

In our example, treatment regimes refer to decision rules that describe how Algebra I should be assigned as a function of individual characteristics. Among various decision rules, we are interested in identifying OTRs that maximize the expected achievement score in 11th grade, by leveraging the heterogeneous effects resulting from individual characteristics. Specifically, we use the heterogeneous effects of Algebra I based on prior math achievement and math interest levels to identify those who would benefit from taking

Algebra I in 9th grade.

To formalize, suppose that we have a decision point $M$, where there are two options as $\{0, 1\} \in \mathcal{M}$. Let $h \in \mathcal{H}$ denote history variables available on an individual at decision point $M$. Given our DAG in Figure 1, history variables are $H = (R, X, C)$. Consider a decision rule for $M$ based on history variables, which is denoted as $d(H) \in \mathcal{D}$, where $\mathcal{D}$ is all possible decision rules. For example,
$d(H) = I(\texttt{math score} > -0.5 \; \& \; \texttt{math interest} > 0.3)$, indicates that the Algebra I course will be recommended to those whose previous math score is greater than -0.5 and whose interest level is greater than 0.3. Of these decision rules, our primary interest is to identify optimal decision rules $d^{opt} \equiv d^{opt}(H) \in \mathcal{D}$. Assuming that larger outcomes are preferred, $d^{opt}$ is the decision rules that maximize the value function $V(d)$, which is an potential outcome under the optimal decision. Formally, OTRs are expressed as (Tsiatis, Davidian, Holloway, & Laber, 2019):

$$d^{opt} = \arg\max_{d \in \mathcal{D}} V(d) = \arg\max_{d \in \mathcal{D}} E\{Y(d)\}. \quad (1)$$

Although there are many estimation methods to obtain these optimal rules, we will briefly review the two basic estimators that we used in our analysis: Q-Learning and weighting.

**Q-Learning.** One intuitive method to obtain optimal decision rules is Q-Learning (e.g., Qian & Murphy, 2011). This approach involves fitting a model for the Q-function, which represents the conditional expectation of the outcome given the risk factor and history variables as $Q(h, m) = E[Y|H = h, M = m]$. However, the true Q-function is unknown, and hence must be estimated from the data. Therefore, we assume that the Q-function follows a linear parametric model as:

$$Q(H, M; \beta) = \beta_0 + \beta_1 H + (\beta_2 + \beta_3 H_1) \times M, \quad (2)$$

where $H_1 \subset H$. Here, $H$ contain history variables, while $H_1$ represents variables that have heterogeneous (interaction) effects on the outcome based on the risk factor $M$. Identifying a subset of variables with heterogeneous effects ($H_1$) is crucial and should be grounded in substantive understanding. For example, in the case of high school math courses, students lacking interest or prior knowledge may not experience the same benefits. In this scenario, interest or prior knowledge can serve as elements of $H_1$. By leveraging these heterogeneous effects, we can construct an optimal decision for $M$ as $d^{opt} = I(\hat{\beta}_2 + \hat{\beta}_3 H_1 > 0)$. If optimal decision rules are followed, the expected outcome will be maximized. Q-Learning is intuitive and easy to understand. However, the quality of Q-Learning depends on the correct specification of the outcome model, which is often violated.

**Weighting.** To mitigate the risk of the outcome misspecification, a weighting method was proposed by B. Zhang, Tsiatis, Davidian, Zhang, and Laber (2012). The weighting method requires specifying a model for the risk factor without the need to fully specify the outcome model, making it more robust to potential misspecification of the outcome model. However, it relies on the correct specification of the model for the risk factor. To maximize the value function, it first requires constructing the following contrast function:

$$C(Y, M, H) = \frac{MY}{P(M|H;\hat{\gamma})} - \frac{(1-M)Y}{1-P(M|H;\hat{\gamma})}, \tag{3}$$

where $\hat{\gamma}$ represents the regression coefficient estimates in the propensity model. For example, this contrast function estimates the difference in potential math score in 11th grade between those who take Algebra I in 9th grade and those who do not. Then, we can define $Z = I(C(Y, M, H) > 0)$, so that $Z = 1$ indicates subjects who will benefit from $M = 1$ than $M = 0$.

The next step is to find estimated OTRs using classification techniques. Based on the estimated contrast function, the optimal rule is obtained by minimizing a weighted

classification error as

$$d^{opt}(H) = \arg\min_{d\in\mathcal{D}} \sum |C(Y,M,H)| \left[\hat{Z} - d(H)\right]^2. \tag{4}$$

The optimal decision rules are obtained by minimizing the difference between $\hat{Z}$ and the decision rules expressed as a function of $H$. This optimization problem can be solved by existing classification techniques, such as classification and regression trees (CART, Breiman, 2017). CART avoids making a specific functional form assumption and is relatively robust to model misspecification (Setoguchi, Schneeweiss, Brookhart, Glynn, & Cook, 2008). This approach is also flexible, as it separates maximizing the value function and estimation of the contrast function. In the weighting estimator, $H_1$ is either explicitly specified as $d(H_1)$ or learned through CART during the optimization process. A comprehensive overview of both Q-Learning and weighting methods can be found in Chapter 3 of Tsiatis et al. (2019).

**An Application to HSLS:09.** To identify optimal decision rules for whether students should take Algebra I in 9th grade ($M$), we considered students' math efficacy, students' interest in math courses, and their grades in the most advanced 8th-grade math course ($H_1$). While it is possible that the effect of taking Algebra I may vary by school context, we believe it is inappropriate to recommend that students delay Algebra I due to poor school quality. Doing so could inadvertently reinforce inequities. Therefore, we did not consider school context for $H_1$. Confounders $X_1, X_2$ and $C$ were used as main effects in the Q-Learning approach and for calculating propensity score in the weighting approach. We fitted separate models for Black and White students to ensure that optimal decisions were not disproportionately influenced by the majority student group. However, this approach results in applying different criteria in recommending Algebra I by 9th grade, which may prompt discussions about equal opportunity. While this topic is important, it lies beyond the scope of this paper. For further discussion, refer to Kamiran and Calders (2012) and Hardt, Price, and Srebro (2016).

Table 1 presents the percentage of students who will be recommended to take Algebra I in 9th grade and the percentage of students whose decisions aligned with the optimal decision rules (i.e., compliance) in the observed data. According to optimal rules, more than 95% of students would be recommended to enroll in Algebra I in 9th grade, with Q-Learning suggesting 95.6% and the weighting method indicating 98.1%.

Specifically, the Q-Learning method recommends Algebra I for Black students who satisfy this condition:
$I(2.066 \times \texttt{grade} + 1.465 \times \texttt{math efficacy} + 0.845 \times \texttt{math interest} > 0.085)$. Similarly, it recommends Algebra I for White students who satisfy this condition:
$I(0.121 \times \texttt{grade} + 0.090 \times \texttt{math efficacy} - 0.062 \times \texttt{math interest} > 0.157)$. According to this rule, similar proportions of students within each race would be recommended to take Algebra I (Black: 94.8% vs. White: 95.9%), maximizing the average scores at 0.054 for Black students and .460 for White students.

Meanwhile, the weighting method recommends Algebra I for 100 % of White students but only for 90.0% of Black students who have either a math interest score of at least 1 SD or a grade of A, maximizing the average scores at 0.069 for Black students and 0.459 for White students. Although fewer Black students were recommended based on the weighting method, the expected score is higher than that of Q-Learning (0.069 with the weighting method vs. 0.054 with Q-Learning). This disproportionate recommendation across racial groups is likely due, in part, to Black students facing greater barriers in achieving the necessary level of prior math achievement and interest to benefit from taking Algebra I by 9th grade.

The next question is: what percentage of students' enrollment decisions aligned with the recommendations obtained from the optimal decision rule? Interestingly, both Q-Learning and weighting methods reveal that enrollment decision patterns of White students align more closely with the optimal decision rules compared to those of Black students. Specifically, 80.4% of White students' decisions are consistent with the Q-Learning recommendations, compared to 76.4% for Black students. Similarly, 83.0% of White students' decisions are consistent with the recommendations derived from the weighting method, whereas 74.0% of Black students' decisions align with these recommendations. This suggests that White students' enrollment decision patterns are more aligned with the rules that maximize their achievement in 11th grade compared to Black students, even when OTRs were determined separately for each group. This finding potentially indicates that fewer Black students are in circumstances that allow them to enroll in Algebra I by 9th grade, despite the potential benefits of taking the course at that time.

Hereafter, we use the term "compliance" to refer to whether a student's decision aligns with the recommendation. Formally, $I(M = d^{opt}) = 1$ where $I(\cdot)$ is an indicator function. This compliance is determined by the observed consistency between the student's decision and the recommendation.

Table 1

*Recommendation and compliance rates by race*

| | Recommendation (%) | | | Compliance (%) | | |
|---|---|---|---|---|---|---|
| | Black | White | Total | Black | White | Total |
| Q-Learn | 94.8 | 95.9 | 95.6 | 76.4 | 80.4 | 79.6 |
| Weighting | 90.9 | 100 | 98.1 | 74.0 | 83.0 | 81.3 |

Note. Recommendation: Percentage of students who would be recommended to take Algebra I in 9th grade, Compliance: Percentage of students whose course taking pattern aligns with the optimal rule. Source: U.S. Department of Education, Institute of Education Sciences, National Center for Education Statistics. High School Longitudinal Study of 2009 (HSLS:09) Base-Year Restricted-Use File (NCES 2011-333)

### 3.2. Causal Decomposition Analysis with Individualized Interventions

Causal decomposition analysis begins by estimating the initial disparity, defined as the average difference in an outcome between groups within the same levels of outcome-allowable covariates. Formally,

$$\tau_c \equiv E[Y|R=1,c] - E[Y|R=0,c], \text{ for } c \in \mathcal{C} \quad (5)$$

where $R = 1$ is the comparison group (Black students) and $R = 0$ is the reference group (White students). For simplicity, we focus on two groups in this manuscript; however, all the subsequent definitions can be generalized to accommodate multiple comparison groups. Causal decomposition does not aim to estimate the causal effect of socially ascribed characteristics such as race/ethnicity or gender, as these are non-modifiable. Instead, it focuses on estimating the causal effect of malleable risk factors (VanderWeele & Robinson, 2014).

In the example, the initial disparity represents the average difference in math achievement in the 11th grade between Black and White students, within the same gender group. We controlled for gender since males are slightly overrepresented in each racial group (51.3% for Whites and 53.1% for Blacks), and adjusting for these differences help address potential biases due to restricting data to students who did not take Algebra I before 9th grade. Additionally, gender is a factor contributing to outcome differences that we consider "allowable" to remove when defining racial disparities. Although we refer to them as allowable covariates, they can include a subset of both baseline covariates ($C$) and intermediate confounders ($X$). For a detailed discussion on allowability, refer to Jackson (2021).

**Individualized Controlled Direct Effects (ICDE).** Park et al. (2024a) proposed an intervention in which all students followed optimal decision rules based on their prior achievement and interest levels. The recommendation obtained from OTRs serve as a reference for each student, and we can examine whether this hypothetical intervention reduces disparities in an outcome. Formally, disparity remaining at $M = d^{opt}$ is defined as

$$\zeta_c^{\texttt{ICDE}}(d^{opt}) \equiv E[Y(d^{opt})|R=1,c] - E[Y(d^{opt})|R=0,c], \text{ for } c \in \mathcal{C} \quad (6)$$

where $d^{opt}$ is an optimal value for risk factor $M$. This definition of disparity remaining states the difference in an outcome between the comparison (e.g. Black students) and reference groups (e.g. White students) after setting their risk factor (Algebra I) to the optimal value obtained from OTRs. Park et al. (2024a) referred to this quantity as *individualized controlled direct effects* (ICDE) given that the risk factor is fixed to a pre-specified value–i.e., optimal values obtained in response to individual characteristics.

Given assumptions A1-A3, the ICDE can be estimated by fitting a marginal structural model with baseline covariates centered at $C = c$ as:

$$Y = \gamma_1 + \gamma_2 R + \gamma_3 C + \epsilon_2, \tag{7}$$

given the weight of $W = I(M = d^{opt})/P(M|R, X, C)$. The disparity remaining is then estimated as $\hat{\zeta}_c^{\texttt{ICDE}}(d^{opt}) = \hat{\gamma}_2$. If significant, the interaction effects between $R$ and $C$ can be specified. In this case, the disparity remaining is still given by $\hat{\zeta}_c^{\texttt{ICDE}}(d^{opt}) = \hat{\gamma}_2$.

**Individualized Interventional Effects (IIE).** However, requiring all students to adhere strictly to optimal decision rules is neither realistic nor necessarily beneficial. Some students may have valid reasons to take Algebra I in 9th grade, even if it was not recommended for them. A more feasible approach is to ensure that Black students comply with recommendations at the same rate as White students among individuals who share the same levels of target-factor-allowable covariates (denoted as $A^m$). Equalizing compliance with recommendations within the same levels of allowable covariates requires an equity-based judgment–determining which variables should be considered fair or appropriate for assigning interventions at the same level. For instance, it may not be fair to equalize compliance within the same levels of student SES, as student SES is a source of structural inequality. In our example, we did not specify any covariates as target-factor-allowable covariates for the fairness of the intervention.

To precisely define, let

$K = G_{I(M=d^{opt})|R=0,A^m} \times d^{opt} + (1 - G_{I(M=d^{opt})|R=0,A^m}) \times (1 - d^{opt})$. Then, disparity reduction and disparity remaining due to equalizing compliance rates across groups can be expressed as follows:

$$\begin{aligned} \delta_c^{\texttt{IIE}}(1) \equiv& E[Y|R=1,c] - E[Y(K)|R=1,c], \text{ and} \\ \zeta_c^{\texttt{IIE}}(0) \equiv& E[Y(K)|R=1,c] - E[Y|R=0,c], \end{aligned} \tag{8}$$

where $d^{opt}$ is an optimal value for risk factor $M$, and $G_{I(M=d^{opt})|R=0,A^m}$ is a random draw from the compliance distribution for $M$ of the reference group given target-factor-allowable covariates $A^m$. The notation $Y(G_{I(M=d^{opt})|R=0,A^m} \times d^{opt})$ indicates a potential outcome under the value of $M$ that is determined by a random draw from the compliance distribution of the reference group among individuals with the same levels of target-factor-allowable covariates. For example, if a random draw indicates that a reference group individual complied with the optimal value, the corresponding individual in the comparison group will likewise adopt the optimal value. Disparity reduction ($\delta_c^{\texttt{IIE}}(1)$) represents the disparity in outcomes among a comparison group (e.g., Black students) after intervening to setting the compliance rate equal to that of a reference group (e.g., White students) within the same target-factor-allowable covariate levels. Disparity remaining ($\zeta_c^{\texttt{IIE}}(0)$) quantifies the outcome difference that persists between a comparison and reference group even after the intervention.

Under assumptions A1-A3, the calculation of disparity reduction and disparity remaining using a regression estimator follows these steps. First, fit a compliance model as:

$$P(I(M = d^{opt}) = 1|R, A^m) = \text{logit}^{-1}(\phi_1 + \phi_2 R + \phi_3 A^m), \tag{9}$$

where $I(\cdot)$ is an indicator function. Next, fit a marginal structural model using centered baseline covariates $C = c$, formulated as:

$$Y = \lambda_1 + \lambda_2 R + \lambda_3 I(M = d^{opt}) + \lambda_4 C + \epsilon_3 \tag{10}$$

given the weight of $W = P(M|R, X, C)^{-1}$. Then, $\hat{\delta}_c^{\texttt{IIE}}(1)$ can be obtained as $\left\{\frac{\exp(\hat{\phi}_1+\hat{\phi}_2+\hat{\phi}_3\hat{E}[A^m])}{1+\exp(\hat{\phi}_1+\hat{\phi}_2+\hat{\phi}_3\hat{E}[A^m])} - \frac{\exp(\hat{\phi}_1+\hat{\phi}_3\hat{E}[A^m])}{1+\exp(\hat{\phi}_1+\hat{\phi}_3\hat{E}[A^m])}\right\} \times \hat{\lambda}_3$. In cases where an interaction exists between race $R$ and compliance $I(M = d^{opt})$, $\hat{\delta}_c^{\texttt{IIE}}(1)$ can be expressed as $\left\{\frac{\exp(\hat{\phi}_1+\hat{\phi}_2+\hat{\phi}_3\hat{E}[A^m])}{1+\exp(\hat{\phi}_1+\hat{\phi}_2+\hat{\phi}_3\hat{E}[A^m])} - \frac{\exp(\hat{\phi}_1+\hat{\phi}_3\hat{E}[A^m])}{1+\exp(\hat{\phi}_1+\hat{\phi}_3\hat{E}[A^m])}\right\} \times (\hat{\lambda}_3 + \hat{\lambda}_5)$ where $\lambda_5$ represents the coefficient for the interaction effect, and $\hat{\zeta}_c^{\texttt{IIE}}(0)$ can be obtained as $\hat{\tau}_c - \hat{\delta}_c^{\texttt{IIE}}(1)$

Alternatively, a weighting estimator can be used to estimate disparity reduction and disparity remaining through the following steps. First, compute the compliance rate among the reference group $R = 0$ given $A^m$, denoted as $\pi_{I=\theta|0,A^m} \equiv P[I(M = d^{opt}) = \theta|R = 0, A^m]$. For each value of $\theta$, weights can be formulated as $W_{\texttt{IIE}}^{\theta} \equiv \frac{I\left(M=\theta\times d^{opt}+(1-\theta)\times(1-d^{opt})\right)}{P(M|R=1,X,C)}$. Finally, the disparity reduction is estimated as $\hat{\delta}_c^{\texttt{IIE}}(1) = \hat{E}[Y|R = 1, C = c] - \sum_{\theta} \hat{\pi}_{I=\theta|0,A^m} \hat{E}[\hat{W}_{\texttt{IIE}}^{\theta} Y|R = 1, C = c]$ and disparity remaining is estimated as $\hat{\zeta}_c^{\texttt{IIE}}(0) = \sum_{\theta} \hat{\pi}_{I=\theta|0,A^m} \hat{E}[\hat{W}_{\texttt{IIE}}^{\theta} Y|R = 1, C = c] - \hat{E}[Y|R = 0, C = c]$. To address multiple-stage estimation, we used bootstrapping to estimate the standard errors of disparity remaining and disparity reduction.

**An Application to HSLS:09.** For subsequent analyses, we use the optimal decision rule obtained from the weighting method, which is relatively robust to misspecification of the outcome model. We estimate initial disparity, disparity remaining, and disparity reduction with ICDE and IIE as described in Sections 3.2 and 3.3. Table 2 presents estimates of quantities of interest. The initial disparity in 11th grade math achievement between Black and White students is -0.413 SD and significant at the $\alpha = 0.001$ level, indicating that Black students achieve significantly lower scores than White students, controlling for gender.

We then estimate the remaining disparity using ICDE. Forcing every students to follow the optimal rule obtained from the weighting method results in a remaining disparity of $\zeta_c^{\texttt{ICDE}} = -0.477$ SD, representing a 15.5% increase from the initial disparity (a widening gap between racial groups). This larger disparity indicates that while both White and Black students benefit from following optimal rules in terms of maximizing their scores, White students experience a greater improvement than Black students.

Equalizing the proportion of students whose decisions align with the recommendations between groups results in a remaining disparity of $\zeta_c^{\texttt{IIE}} = -0.395$ SD, which represents a 4.4% reduction from the initial disparity. In the analyses, we have incorporated the interaction between race and compliance. Our findings support minorities' diminished return hypothesis (Assari, 2020), showing that the effect of compliance was greater for White students than for Black students. Consequently, after incorporating the interaction effect, the disparity reduction due to equalizing compliance rates across groups is minimal (4.36%) and not statistically significant.

Table 2

*Estimates of the initial disparity, disparity reduction, and disparity remaining*

| | Estimate (S.E.) | |
|---|---|---|
| | ICDE | IIE |
| Initial disparity | -0.413*** (0.010) | -0.413*** (0.010) |
| Disparity remaining | -0.477*** (0.059) | -0.395*** (0.035) |
| Disparity reduction | | -0.017 (0.013) |
| % reduction | -15.5% | 4.36% |

Note. ICDE: Individualized controlled direct effect, IIE: Individualized interventional effect. The asterisk followed by estimates indicates the level of statistical significance (*: significant at 0.05, **: at 0.01, ***: at 0.001). Gender and native language are centered at the mean. The regression estimator was used for the IIE estimates. Source: U.S. Department of Education, Institute of Education Sciences, National Center for Education Statistics. High School Longitudinal Study of 2009 (HSLS:09) Base-Year Restricted-Use File (NCES 2011-333)

These results suggest that while following optimal rules as well as equalizing the compliance rate with the optimal rule between the groups may maximize the average math achievement at 11th grade, it does not effectively reduce the disparity in math achievement.

These findings can only be interpreted causally if assumptions A1–A3 are met. Therefore, in the next section, we develop a sensitivity analysis to evaluate the robustness of the findings to potential violations of the no omitted confounding assumption A1.

## 4. Simulation-based Sensitivity Analysis for Causal Decomposition

The goal of sensitivity analysis is to understand what the effect estimates would be if we measured and conditioned on omitted confounder $U$. Carnegie et al. (2016) proposed a simulation-based approach to sensitivity analysis to account for the absence of omitted confounding for continuous outcomes. Qin and Yang (2022) subsequently extended this approach to the context of mediation analysis and addressed different types of outcomes and unmeasured confounders. In this section, we extend this simulation-based sensitivity analysis to estimate OTRs as well as disparity reduction and disparity remaining using ICDE and IIE.

### 4.1. Framework

Following Carnegie et al. (2016), we proceed with three steps: 1) specify a model to compute a complete-data likelihood, 2) use this model to derive conditional distribution of unmeasured confounder $U$ on observed variables, and 3) compute disparity reduction or disparity remaining using generated unmeasured confounder $U$ and sensitivity parameters.

First, a complete-data likelihood can be expressed as:

$$P(Y, M, U|R, X, C) = P(Y|R, X, U, M, C)P(M|R, X, U, C)P(U|R, X, C). \tag{11}$$

This factoring allows the sensitivity parameters to be coefficients of unmeasured confounder $U$ on the risk factor $M$ and the outcome $Y$, which is straightforward to interpret for applied researchers. Previous studies assume the independence of the unmeasured confounder $U$ with the remaining confounders $R$, $X$, and $C$ ($U \perp R, X, C$) to reduce the number of sensitivity parameters. In disparities research, unmeasured confounders are likely affected by the group status $R$ and demographic variables $C$. Therefore, we conceptualize unmeasured confounder $U$ as the remaining portion after accounting for $R, X$, and $C$.

To calculate the complete-data likelihood, we specify the following models for the outcome and the intervening factor. We assume a normal distribution for the outcome, and a Bernoulli distribution for the binary risk factor, consistent with our example.

$$\begin{aligned} &Y|R, X, U, M, C \sim N(\beta_r^y R + \beta_x^y X + \beta_u^y U + \beta_m^y M + \beta_{mh_1}^y M H_1 + \beta_c^y C, \sigma^2_{y|rxumc}), \text{ and} \\ &M|R, X, U, C \sim \text{Bernoulli}\Big(\text{logit}^{-1}(\beta_r^m R + \beta_x^m X + \beta_u^m U + \beta_c^m C)\Big), \end{aligned} \tag{12}$$

where $N(\cdot)$ denotes the normal distribution, Bernoulli$(\cdot)$ denotes the Bernoulli distribution, and $H_1$ is a subset of history variables $H$ that modify the effect of the risk factor (e.g., prior math achievement and math interest). Here, we only included interaction terms between $M$ and $H_1$ for the outcome model, as specified in equation (2). While additional nonlinear terms can be specified as needed, caution is advised when interaction terms involving $U$ are included. Refer to Section 4.2 for performance results of this simulation-based sensitivity analysis in the presence of such interaction terms.

For unmeasured confounder $U$, we assume a normal distribution when $U$ is continuous and a Bernoulli distribution when $U$ is binary:

$$U|R, X, C \sim \begin{cases} N(0, \sigma^2_{u|rxc}) \text{ for a continous } U, \\ \text{Bernoulli}(\pi) \text{ for a binary } U, \end{cases}$$

where $\sigma^2_{u|rxc}$ is the variance of $U$, and $\pi$ is the proportion of $U = 1$. We assume that the continuous $U$ is centered at zero and for $\sigma^2_{u|rxc}$, we use an existing covariate as a benchmark, which will be explained in Section 5.1.

The sensitivity parameters for the outcome and the risk factor are denoted as $\beta_u^y$ and $\beta_u^m$, respectively. The sensitivity parameters represent the effect of the unmeasured confounder $U$ on the outcome and the risk factor. The purpose of sensitivity analysis is to identify plausible combinations of these sensitivity parameters that could potentially alter the study's conclusions or affect the significance level.

Second, the conditional probability of $U$, $P(U = 1|Y, M, R, X, C)$, can be derived using the formula below. For a binary $U$, we have

$$P(U = 1|Y, M, R, X, C) = \frac{f(Y|R, X, M, C, U = 1)P(M|R, X, C, U = 1)P(U = 1)}{\sum_u f(Y|R, X, M, C, U = u)P(M|R, X, C, U = u)P(U = u)}, \tag{13}$$

where $u \in \{0, 1\}$. When $U$ is continuous, the denominator should be integrated over the values of $u \in \mathcal{U}$. When $U, M$, and $Y$ are all continuous, a closed form expression of the distribution of $U$ conditional on observed variables can be obtained. For the derivation and its result, refer to Carnegie et al. (2016) and Qin and Yang (2022). For a binary $M$, the stochastic EM algorithm (Feodor Nielsen, 2000) was used to estimate the unknown parameters $\theta = (\beta^m_r, \beta^m_x, \beta^m_c, \beta^y_r, \beta^y_x, \beta^y_m, \beta^y_{mh_1}, \beta^y_c)$, which consists of two steps: 1) E-step: simulate unmeasured confounder $U$ from its conditional distribution, given the parameters obtained in the previous iteration and the specified sensitivity parameters, and 2) M-step: maximizing an expected complete data log-likelihood with respect to parameters. Then, these two steps are iterated until convergence.

Finally, once the EM algorithm converges, we first obtain OTRs and then estimate the disparity reduction and disparity remaining conditional on the generated $U$ and specified sensitivity parameters. Note that a certain degree of uncertainty is involved when generating the unmeasured confounder $U$ from its conditional distribution. To address this uncertainty, previous studies suggest running the EM algorithm multiple times (e.g., 30 times) and repeatedly estimating the effects of interest given the multiple values of $U$. The estimates are then combined by averaging over the number of simulations, and standard errors are computed based on Rubin's rule (Rubin, 2004).

While this approach effectively addresses the uncertainty related to generating $U$, it is computationally intensive especially when $U$ is continuous where `integrate()` R function is used. Therefore, we made a minor adjustment to the existing algorithm to simulate multiple values of $U$ based on its conditional distribution without repeating the `integrate()` R function.

### 4.2. Simulation Study

The purpose of this simulation study is to evaluate the performance of simulation-based sensitivity analysis in identifying OTRs and in estimating disparity reduction and disparity remaining with individualized interventions in the presence of unmeasured confounder $U$. Specifically, we evaluate the accuracy in obtaining OTRs and the performance in terms of bias and variance of estimating the effects of interest, comparing with and without adjustment for unmeasured confounder $U$. The evaluation is conducted under two scenarios: when $U$ exerts constant effects, and when $U$ exhibits heterogeneous effects with the risk factor. Additionally, we vary the sample sizes (500, 1000, and 2000) and the magnitude of sensitivity parameters (0.5, 1, and 1.5).

***Data Generation and Simulation Setting.***

To generate the population-level simulation data, we create the baseline covariate $C$, the social group $R$, intermediate confounders $X_1$, $X_2$, $X_3$, the unmeasured confounder $U$, and the outcome variable $Y$ as follows with a population size of $10^6$. Here, $X_1$, $X_2$, $X_3$, and $Y$ are generated as continuous variables, while the remaining variables are binary, taking values of 0 or 1. Specifically, we generate a binary covariate $C$ with a probability of 0.4 for $C = 1$ and a binary social group status $R$ with the probability $\text{logit}^{-1}(1 - 0.5C)$ for $R = 1$. We generate an unmeasured confounder $U$ with a probability of 0.5 for $U = 1$, as we conceptualize $U$ as the remaining part after accounting for $R, X$, and $C$. The $X$ variables are generated as follows:

$$X_1 = -0.8 + R + 1.5C + \epsilon_1$$
$$X_2 = 0.5 + 0.5R + 0.5C + \epsilon_2$$
$$X_3 = -0.8 - R + 0.5C + \epsilon_3,$$

where the error terms $\epsilon_1$, $\epsilon_2$, and $\epsilon_3$ are drawn from a standard normal distribution. When $U$ has constant effects, the optimal decision rule dictates that $M$ should take 1 for subjects with $X_1 > 0.1$ and $X_2 > 0.1$. When $U$ has heterogeneous effects, the optimal decision rule dictates that $M$ should take 1 for subjects with $X_1 > 0.1$ and $U > 0.5$. However, subjects determine the value of $M$ according to the following logit function: $\text{logit}^{-1}(0.5 - 0.5R + 0.2X_1 + 0.5C + \beta_u^m U)$, where $\beta_u^m$ is a pre-set true value of the sensitivity parameter. Although it is challenging to generate the exact form that maximizes the outcome under the OTR, it can be approximated. Following Z. Zhang (2019), the outcome $Y$ is generated as:

$$Y = 0.5 - 0.5R + 0.25X_1 + 0.25X_2 - 0.25X_3 - \beta_u^m(M - M_{\text{opt}})^2 + 0.25C + \beta_u^y U + \epsilon_4$$

where $\epsilon_4$ is drawn from a standard normal distribution, and $\beta_u^y$ is a pre-set true value of the sensitivity parameter.

In this study, the population-level simulation data is generated using varying magnitudes of sensitivity parameters $(\beta_u^y, \beta_u^m) \in \{(0.5, 0.5), (1, 1), (1.5, 1.5)\}$. After generating the data, we randomly selected subsets with sample sizes of $n \in \{500, 1000, 2000\}$. The proposed simulation-based sensitivity analysis described in Section 4.1 was then conducted. For the homogeneous effect of $U$, the outcome model specified in equation (12) included interaction terms between $M$ and $X_1$ as well as $M$ and $X_2$. For the heterogeneous effect of $U$, the outcome model included interaction terms between $M$ and $X_1$ as well as $M$ and $U$. The simulation results are based on 500 iterations.

***Simulation Results.*** Figure 2 presents the accuracy of predicting OTRs, with the maximum possible accuracy of 1 indicated by red dashed lines. When $U$ exerts constant effects, OTR predictions are relatively robust to omitted variable bias. With sensitivity parameters of 0.5 and 1, the accuracy exceeds 0.95 with a sample size of 2000, both with adjusting for $U$ (green boxplots in Figure 2) and without adjusting for $U$ (red boxplots). However, when the sensitivity parameter is 1.5, accuracy generally improves after adjusting $U$. For example, the median accuracy increases from 0.74 before adjustment to 0.78 after adjustment for a sample size of 500.

When $U$ exerts heterogeneous effects, the accuracy of predicting OTRs substantially improves after adjusting for $U$ across all sample sizes and sensitivity parameters. However, the accuracy does not reach the same level as when $U$ exerts constant effects. This improvement is most pronounced with a sample size of 2000 and sensitivity parameters of 1.5, where the median accuracy increases from 0.69 before adjustment to 0.81 afterward.

In Figure 3, we present boxplots of the estimated values of $\zeta_c^{\texttt{ICDE}}$, $\delta_c^{\texttt{IIE}}$ using the regression estimator, and $\delta_c^{\texttt{IIE}}$ using the weighting estimator, with their true values represented by red dashed lines. The results for $\zeta_c^{\texttt{IIE}}$ using both the regression and weighting estimators are provided in Appendix A in the online supplement, and show little difference compared to $\delta_c^{\texttt{IIE}}$ using the respective regression and weighting estimators.

When $U$ exerts constant effects, the unadjusted estimates (green boxplots in Figure 3) remain relatively robust with a sensitivity parameter of 0.5. However, as the magnitude of the sensitivity parameters increases, the unadjusted estimates become increasingly biased. In contrast, the adjusted estimates remain accurately centered around the true value, regardless of the sensitivity parameters, particularly as the sample size increases. For $\delta_c^{\texttt{IIE}}$, the regression estimator performs slightly better in terms of bias compared to the weighting estimator.

When $U$ exerts heterogeneous effects, the adjusted estimates show substantial improvement in terms of bias compared to the unadjusted estimates. However, despite this improvement, $\delta_c^{\texttt{IIE}}$ using the regression estimator remain biased, even with a sample size of 2000. In contrast, $\zeta_c^{\texttt{ICDE}}$ and $\delta_c^{\texttt{IIE}}$ using the weighting estimator are centered around the true value as the sample size increases.

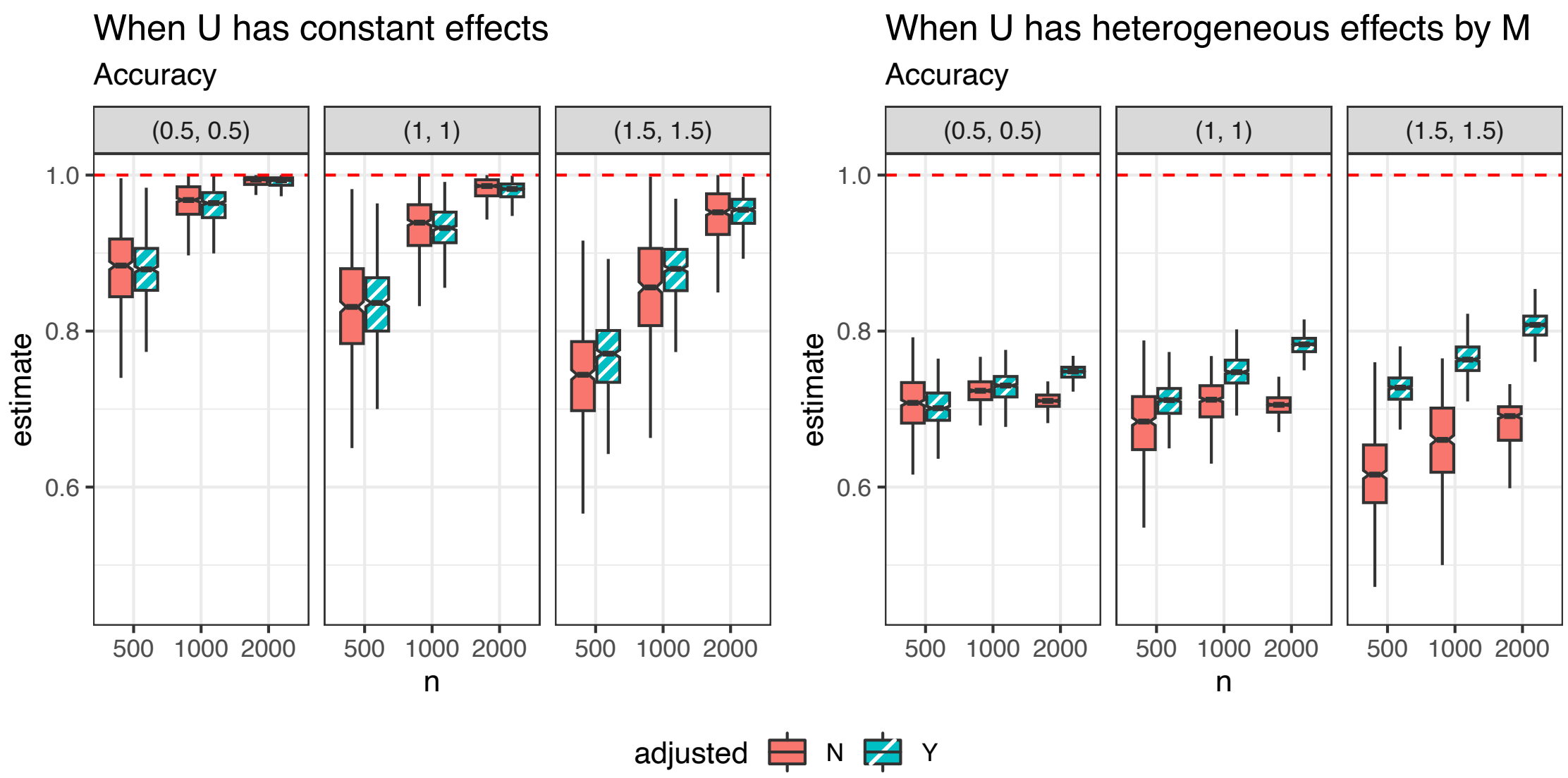

*Figure 2*. Accuracy of the optimal value when $U$ has constant (left) and heterogeneous (right) effects by $M$. In each panel, the red dashed line represents an accuracy of 1. 'N' refers to not adjusted for $U$, and 'Y' refers to adjusted for $U$.

Overall, our simulation study results indicate that the proposed sensitivity analysis is effective when $U$ has constant effects, addressing two sources of biases arising from obtaining OTRs and from estimating disparity reduction and disparity remaining. However, When $U$ exhibits heterogeneous effects across levels of the risk factor $M$, the accuracy of predicting OTRs decreases to about 0.8, even with a sample size of 2000, and the regression-based estimator for IIEs become biased. This decline in accuracy of predicting OTRs is due to the misspecification of the outcome model used to simulate the unmeasured confounder $U$ (see equation (12)). The regression-based estimator of IIEs is particularly susceptible to this outcome misspecification, as it relies on the specified outcome model shown in equation (10). The implications of this reduced performance is further discussed in Section 6.

As a sensitivity analysis, we also examined the performance when the distribution of $U$ is skewed. Appendix B in the online supplement provides results using left- and right-skewed distributions of $U$, showing that overall performance trends remain similar to those in the symmetric case. While the bias was slightly larger in some cases for skewed distributions of $U$, the increased bias does not affect inference significantly. The 95% confidence interval coverage rates indicate that adjusting for $U$ maintains coverage near or above the nominal level (0.95). In contrast, failing to adjust for $U$ leads to lower coverage, highlighting biases introduced by unaccounted confounding.

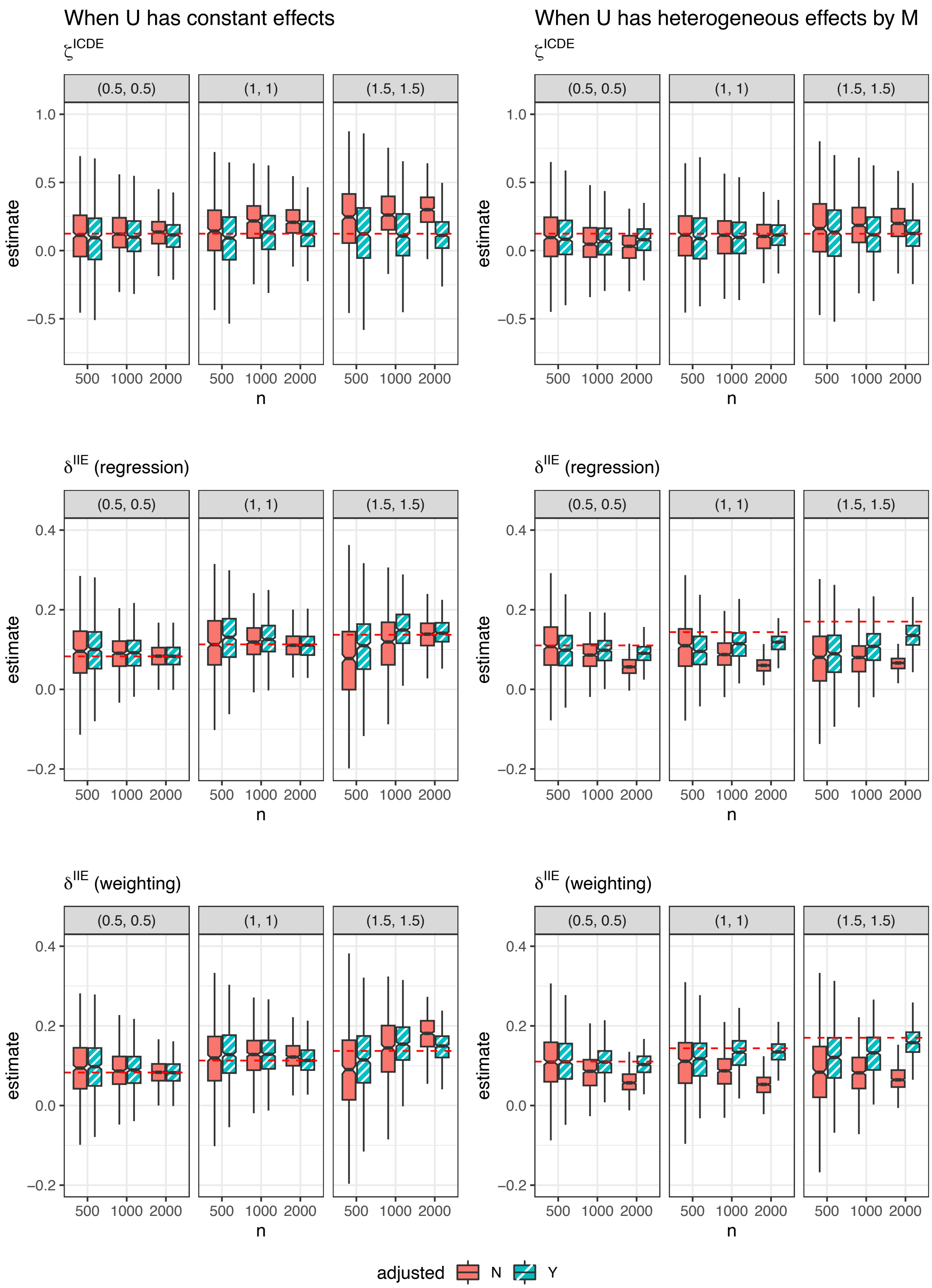

*Figure 3*. Estimates of $\zeta^{\texttt{IIE}}$ using the regression estimator and $\zeta^{\texttt{IIE}}$ using the weighting estimator, when $U$ has constant (left) and heterogeneous (right) effects by $M$. In each panel, the red dashed line represents the true value of the estimates. 'N' refers to not adjusted for $U$, and 'Y' refers to adjusted for $U$.

It is noteworthy that the 95% confidence interval coverage rates for ICDE are approximately at the nominal level. However, the coverage rates for IIEs, whether estimated using regression-based and weighting-based methods, exceed the nominal level, indicating that the standard errors are larger, leading to conservative inference. The implications of these larger standard errors for IIE estimates are discussed in Section 6.

## 5. Enhancing the Interpretability of Simulation-Based Sensitivity Analysis

Sensitivity analysis is an essential component of causal inference. However, it has been underutilized in the social sciences, perhaps due to its complex nature and difficulty of interpretation (Cinelli & Hazlett, 2020). A crucial question, therefore, is "How can we enhance the interpretability of results obtained from simulation-based sensitivity analysis?" The literature frequently offers graphical tools to illustrate how the estimates and their statistical significance change based on different combinations of sensitivity parameters. However, validating results against potential omitted confounding using graphical tools can be challenging, as estimates can vary from positive to negative and from significant to non-significant, depending on the range of sensitivity parameters.

To make sensitivity analysis more useful, it is important to reduce the range of sensitivity parameters. Previous literature has often employed informal benchmarking strategies using observed covariates, assuming that the strength of unmeasured confounders (represented by sensitivity parameters) would be comparable to the associations between an observed covariate and the outcome, after accounting for all other antecedent variables, except for that observed covariate. However, Cinelli and Hazlett (2020) cautioned that this informal benchmarking strategy could lead to erroneous conclusions due to collider bias (see Section 6 for further discussion). In response, they proposed a formal bounding method to benchmark the strength of unmeasured confounders using observed covariates through $R^2$. However, this approach is not applicable to simulation-based sensitivity analysis when the risk factor is binary.

To address this limitation, we propose a formal benchmarking strategy specifically designed for simulation-based sensitivity analysis when the risk factor is binary. Our covariate benchmark is based on original data scales rather than $R^2$, allowing for more interpretable results with a binary risk factor. It is important to note that the scope of this formal benchmarking strategy is limited to cases with a normally distributed outcome and a binary risk factor. Additionally, this strategy assumes that the unmeasured confounder $U$ is continuous. While binary $U$ offers advantages in terms of computation time, it is more reasonable to assume that $U$ represents a linear combination of several omitted factors, given that researchers usually do not have precise knowledge of unmeasured confounders. In our example, $U$ includes a linear combination of neighborhood factors that are related to the decisions to take Algebra I in 9th grade and math achievement in 11th grade.

### 5.1. Benchmarking With Observed Covariates

In this section, we propose a method to use existing covariates to benchmark the strength of unmeasured confounders using original data scales. We employ an approach that compares the coefficient of the omitted confounder $U$ with that of an observed covariate $X_j$, preferably a significant one, after controlling for the remaining observed covariates $R, X_{-j}$ and $C$. Although we compare the coefficients, we are essentially comparing the strength of the omitted confounder $U$ with that of an observed covariate $X_j$ by assuming equal residual variance between them (i.e., $\sigma^2_{u|rx_{-j}c} = \sigma^2_{x_j|rx_{-j}c}$). In our example, we use childhood SES ($X_j$) for this comparison, as few factors exhibit greater predictive power than childhood SES on both math achievement and the likelihood of taking Algebra I in 9th grade.

To formally compare coefficients of $U$ with that of $X_j$, we define parameters $k_m$ and $k_y$ as follows:

$$k_m := \frac{\exp(\beta^m_{u|rx_{-j}c})}{\exp(\beta^m_{x_j|rx_{-j}c})} \text{ and } k_y := \frac{\beta^y_{u|rx_{-j}c}}{\beta^y_{x_j|rx_{-j}c}}, \tag{14}$$

where $x_{-j} \in \mathcal{X}_{-j}$ represents the vector of intermediate confounders excluding significant variable $X_j$. Here, $\beta^m_{u|rx_{-j}c}$ and $\beta^m_{x_j|rx_{-j}c}$ are the regression coefficient of $U$ and $X_j$, respectively, on $M$ after after conditioning on $R, X_{-j}, C$ in the logit scale; $\beta^y_{u|rx_{-j}c}$ and $\beta^y_{x_j|rx_{-j}c}$ are the regression coefficients of $U$ and $X_j$, respectively, on $Y$ after after conditioning on $R, X_{-j}, C$ in the original $Y$ scale. To enhance interpretability and comparability, we use odds ratio scales for binary $M$ and the original scales for continuous $Y$. Specifically, $k_y$ indicates the extent to which $Y$ is associated with a one unit increase in $U$ relative to how much it is associated with a one unit change in $X_j$, after controlling for $R, X_{-j}$ and $C$. Likewise, $k_m$ indicates the extent to which the odds of $M = 1$ are explained by unmeasured confounder $U$ relative to how much they are explained by $X_j$, after controlling for $R, X_{-j}$ and $C$. These parameters ($k_m$ and $k_y$) should be specified by researchers based on the assumed strength of $U$ relative to that of the given observed covariate $X_j$ (e.g., childhood SES).

To proceed further, we make the following assumptions: B1) the unmeasured confounder $U$ is independent of the remaining covariates $R, X$, and $C$ (i.e., $U \perp R, X, C$) and B2) the effect of $U$ on the outcome as well as the logit scale of the risk factor is constant across the strata of $R, X_{-j}$ and $C$. The first assumption is implied by the fact that $U$ is a remaining part after controlling for $R, X$, and $C$. The second assumption is strong but could be reasonably met in some research contexts.

Under assumptions B1 and B2, we can rewrite our sensitivity parameters given $k_m$ and $k_y$ as below. We use the notation $\beta^m_u = \beta^m_{u|rxc}$ and $\beta^y_u = \beta^m_{u|rxmc}$ to differentiate them with $\beta^m_{u|rx_{-j}c}$ and $\beta^y_{u|rx_{-j}c}$, respectively.

$$\begin{aligned} \beta^m_{u|rxc} &= \ln(k_m) + \beta^m_{x_j|rx_{-j}c}, \text{ and} \\ \beta^y_{u|rxmc} &\lessapprox \left\{ k_y \beta^y_{x_j|rx_{-j}c} - \beta^y_{m|rxc}\pi^m_{u|rxc} \right\} \times \eta, \end{aligned} \tag{15}$$

where $\pi^m_{u|rxc}$ is an expected difference in probability of $M = 1$ when $U$ increases by one unit, after controlling for $R, X$, and $C$; $\eta$ is a function of $k_m$, $\beta^m_{x_j|rx_{-j}c}$, and the conditional variances of $X_j$ ($\sigma_{x_j|rx_{-j}c}$) and $M$ ($\sigma_{m|rxc}$). See Appendix C in the online supplement for details and a proof.

For $k_m = 1$, our sensitivity parameter $\beta^m_{u|rxc}$ is equal to the logit coefficients of the significant observed covariate $X_j$ on the intervening factor $M$, after controlling for the remaining covariates ($\beta^m_{x_j|rx_{-j}c}$). This implies that the informal benchmarking approach–replacing the sensitivity parameter with the coefficient of one observed covariate after removing that covariate from the same controlling set–is valid for $\beta^m_{u|rxc}$ under the assumptions B1 and B2. In contrast, for $k_y = 1$, our sensitivity parameter $\beta^y_{u|rxmc}$ is not equal to the coefficient of the significant observed covariate $X_j$ on the outcome $Y$, after controlling for $R, X_{-j}$ and $C$ ($\beta^y_{x_j|rx_{-j}c}$). Following the informal benchmarking approach, one may wonder whether $\beta^y_{u|rxmc}$ is comparable to the coefficient of the significant observed covariate $X_j$ on the outcome $Y$, after removing that covariate from the same controlling set (i.e., $R, X_{-j}, M, C$). However, this approach also leads to biased result due to collider bias. Refer to Section 6 for a detailed explanation. Overall, this implies that the informal benchmarking approach does not work for $\beta^y_u$ and requires further calculation.

### 5.2. An Application to HSLS:09

We illustrate the newly developed method of benchmarking the strength of unmeasured confounders using our example where the risk factor is binary. Table 3 illustrates the extent to which the disparity reduction and disparity remaining estimates vary if a linear combination of unmeasured confounders was as influential as childhood

SES, or twice as influential as, childhood SES. Positive values of $k_y$ indicate that the unmeasured variables affect the intervening factor and the outcome in the same direction as childhood SES. Conversely, negative values of $k_y$ indicate that the unmeasured variables affect the outcome in a direction opposite to that of childhood SES. In contrast, we consider only positive values of $k_m$, as positive values of $k_m$ indicate that the odds of $M = 1$ explained by the unmeasured confounder $U$ are equal to or greater than the odds explained by childhood SES after controlling the remaining variables.

Table 3

*Estimates of the initial disparity, disparity reduction, and disparity remaining*

| Strength of U relative to SES | Estimate (S.E.) $k_m = 2, k_y = -2$ | $k_m = 1, k_y = -1$ | $k_m = 1, k_y = 1$ | $k_m = 2, k_y = 2$ |
|---|---|---|---|---|
| Initial disparity | -0.413*** (0.010) | -0.413*** (0.010) | -0.413*** (0.010) | -0.413*** (0.010) |
| OTRs | | | | |
| % Recommended | 99.1% | 98.6% | 98.1% | 49.5% |
| ICDE | | | | |
| Disparity remaining ($\zeta_c^{\texttt{ICDE}}$) | -0.436*** (0.075) | -0.454*** (0.059) | -0.453*** (0.059) | -0.453*** (0.066) |
| % reduction | -5.6% | -9.3% | -9.7% | -9.7% |
| IIE | | | | |
| Disparity remaining ($\zeta_c^{\texttt{IIE}}$) | -0.385*** (0.037) | -0.394*** (0.036) | -0.397*** (0.036) | -0.413*** (0.034) |
| Disparity reduction ($\delta_c^{\texttt{IIE}}$) | -0.028 (0.017) | -0.018 (0.014) | -0.016 (0.013) | 0.001 (0.007) |
| % reduction | 6.8% | 4.6% | 3.9% | -0.1% |

Note. 1) ICDE: Individualized Conditional Direct Effects. 2) The asterisk followed by estimates indicates the level of statistical significance (***: at 0.001). 3) The regression estimator was used for the individualized interventional effect (IIE) estimates. Source: U.S. Department of Education, Institute of Education Sciences, National Center for Education Statistics. High School Longitudinal Study of 2009 (HSLS:09) Base-Year Restricted-Use File (NCES 2011-333)

In the context of our example, we assume that unmeasured confounders are as strong as childhood SES or, conservatively, even twice strong as childhood SES. Furthermore, we assume that the unmeasured confounders do not interact with race, intermediate confounders, risk factor, or covariates. Under this condition, we examined how the percentage of students recommended and the effect estimates vary with the level of unmeasured confounding.

First, when the unmeasured confounders are as strong as childhood SES, the percentage of students recommended to take Algebra I in 9th grade remains relatively stable (98.1% when unadjusted v.s. 98.1–98.6% when $k_m = |k_y| = 1$). However, when the unmeasured confounders are twice as strong as childhood SES, the percentage changes significantly, dropping to 49.5% ($k_m = k_y = 2$) or increasing to 99.1% ($k_m = 2, k_y = -2$).

Second, despite this change in the percentage of students recommended, the disparity remaining after following OTRs remains robust. The disparity remaining estimates ($\zeta_c^{\texttt{ICDE}}$) are all significantly negative, ranging from -0.436 ($k_m = 2, k_y = -2$) to -0.454 ($k_m = 1, k_y = -1$). The percentage of disparity reduction ranges from -5.6% to -9.7%, meaning that following OTRs (under the consideration of the unmeasured confounders) would increase the initial disparity.

Third, disparity reduction due to equalizing the compliance rate with optimal rules across groups ($\delta_c^{\texttt{IIE}}$) is statistically insignificant and remains stable in the presence of potential omitted confounders. With the omitted confounding as influential as childhood SES ($k_m = |k_y| = 1$), the disparity reduction estimates ($\delta_c^{\texttt{IIE}}$) remain negative and are not significant, with the estimates ranging from -0.018 to -0.016.

This result was derived under the assumptions of a constant effect of unmeasured confounders, but there could be cases where the effects of unmeasured confounders are heterogeneous. If the unmeasured confounder is school quality and the effect of taking Algebra I is weaker in schools with poor quality, then the sensitivity results obtained would be invalid.

## 6. Discussion

This study presents a sensitivity analysis for OTRs and individualized effects. We summarize our contributions to the literature and highlight opportunities for future research as follows.

**Extending a simulation-based sensitivity analysis to causal decomposition analysis.** The bias in estimating the individualized effects due to unmeasured confounding arises from two sources: 1) identifying OTRs and 2) estimating disparity reduction and disparity remaining due to following OTRs. This bias is not easily addressed through a formula-based sensitivity analysis. Typically, formula-based sensitivity analysis calculates bias as the difference between the estimate and the true effect. While this approach can provide numerical solutions for points where the estimates become zero or lose significance due to confounding, the bias formula must be recomputed for each new estimand. In the case of individualized effects, bias is influenced by a combination of sensitivity parameters and the optimal values, making it challenging to compute.

To address these two sources of bias, we extended a simulation-based sensitivity analysis to assess the robustness of individualized effects. Our simulation study demonstrates that this approach performs well when the unmeasured confounder exerts a constant effect across the strata of remaining variables. However, the accuracy of optimal recommendations declines, and the regression-based estimator for individualized interventional effects (IIEs) is biased when the unmeasured confounder interacts with the risk factor. This reduced performance arises from the misspecification of the outcome model when simulating the unmeasured confounder (see equation (12)). While this issue could be mitigated to some extent if the data-generating model aligns with the specified outcome model, our simulation results indicate that the validity of the simulation-based sensitivity analysis for the regression estimator of IIEs is sensitive to outcome model misspecification. As a result, the weighting estimator is recommended, as it does not rely on outcome modeling. Future research could explore the incorporation of machine learning techniques, which are generally more robust to model misspecification, to enhance the effectiveness of simulation-based sensitivity analyses.

In our case study, we assume a constant effect of unmeasured confounding across the strata of remaining variables. Although this is a strong assumption, applying a sensitivity analysis that accounts for heterogeneous effects of unmeasured confounding may not be practical, particularly for individualized effects. Such an analysis would require more sensitivity parameters than the two currently needed, making it difficult to validate results.

Additionally, optimal decision rules determined by the unmeasured confounder (due to heterogeneous effects of $U$) would complicate meaningful interpretation. For example, recommending Algebra I to students whose unmeasured confounding exceeds 1.5 SD would be meaningless in practice.

Another issue is the larger standard errors for IIEs obtained from the proposed sensitivity analysis. In the simulation study, the proposed method of estimating standard errors tends to be larger across a wide range of generative models, leading to conservative findings. While the literature on optimal treatment regimes recommends the m-out-of-n bootstrap (Shao, 1996; Tsiatis et al., 2019), in our case, this approach resulted in even larger standard errors. Determining the correct standard errors for IIE estimates remain an important area for future research. However, we argue that overcoverage is a less severe issue than undercoverage. Therefore, we recommend using the proposed method despite of its conservative nature until a more refined approach is developed.

**Bounding strategy for binary intervening factors.** Previous literature on sensitivity analysis has often used the association between an observed covariate and the outcome, after accounting for the remaining covariates to benchmark the strength of unmeasured confounding (i.e., $\beta^{y}_{x_j|rx_{-j}mc}$). However, Cinelli and Hazlett (2020) highlighted that this informal benchmark approach may lead to invalid conclusions and demonstrated the potential bias of this approach through a simulation study.

This bias also can be explained by collider bias (Pearl, 2009). A collider is a variable that is influenced by other variables (referred to as 'ancestors') and conditioning on a collider can induce an association between the ancestor variables, even if they are independent to each other. For example, suppose that students with high SES ($X_j$) and those attending high quality school ($U$) are more likely to take Algebra I by 9th grade ($M$). If we know that a student took Algebra I by the 9th grade (that is, conditioning on $M$) and that the student is not from high SES ($X_j$), we can infer that this student is from a high-quality school ($U$). This creates an induced association between $X_j$ and $U$ due to conditioning on $M$. Such collider bias occurs when we use informal benchmarking strategies. When we benchmark the strength of unmeasured confounding against student SES ($X_j$), we use the association between student SES ($X_j$) and the outcome ($Y$) after conditioning on race ($R$), remaining covariates ($X_{-j}$ and $C$), and Algebra I ($M$). We expected that this association captures the direct path between $X_j$ and $Y$, but also inadvertently captures the indirect path between them via $U$, induced by conditioning on $M$.

To avoid this collider bias, the formal covariate benchmark strategy should be used, as suggested by Cinelli and Hazlett (2020). Their method, which is based on $R^2$, can be applied to continuous risk factors and outcomes. However, since OTRs and individualized effects are primarily based on binary risk factors, we developed a formal benchmark method that applies to binary intervening factors using original data scales. While this method was developed in the context of OTRs and individualized effects, it can also be used to estimate any causal effects involving a binary treatment variable.

The newly developed benchmark method relies on the logistic approximation to the cumulative normal distribution. The deviation between the two distributions is less than 0.01 for all values of $u \in \mathcal{U}$ (Ulrich & Wirtz, 2004; Bowling, Khasawneh, Kaewkuekool, & Cho, 2009), making this approximation suitable for practical purposes. Developing a benchmark method for binary treatments without relying on this approximation is a task for future research.

Finally, in extending the simulation-based sensitivity analysis and developing the bounding strategy, we did not account for the multilevel structure where students are nested within schools or neighborhoods. While an alternative approach, such as including school fixed effects in the propensity model and/or the outcome model, could be considered as a quick remedy, this was not feasible in our case study. This is because the outcome (11th-grade math achievement) was measured in high school, while the risk factor (taking Algebra I by 9th grade) was measured in middle school. Addressing the multilevel structure is crucial, and we left this as a direction for future research.

## Declarations

The author(s) declared no potential conflicts of interest with respect to the research, authorship, and/or publication of this article.

## Preregistraction Statement

This study was not preregistered.

## Availability Statement for Data, Code, and Materials

**Data.** For our case study, we used data from the Midlife Development in the U.S. (MIDUS) study. However, as the MIDUS data is restricted from circulation, and the original data can be downloaded from the MIDUS portal at https://www.midus.wisc.edu/data/index.php.

**Code.** The code used during this study and documentation for the code are available in the Github repository, https://github.com/soojinpark33/simulation-sens and have been permanently archived on Zenodo (DOI: 10.5281/zenodo.17038437).

References

Assari, S. (2020). Blacks' diminished health returns of educational attainment: Health and retirement study. *Journal of Medical Research and Innovation*, *4*(2), e000212–e000212.

Bowling, S. R., Khasawneh, M. T., Kaewkuekool, S., & Cho, B. R. (2009). A logistic approximation to the cumulative normal distribution. *Journal of industrial engineering and management*, *2*(1), 114–127.

Breiman, L. (2017). *Classification and regression trees*. Routledge.

Byun, S.-y., Irvin, M. J., & Bell, B. A. (2015). Advanced math course taking: Effects on math achievement and college enrollment. *The Journal of Experimental Education*, *83*(4), 439–468.

Carnegie, N. B., Harada, M., & Hill, J. L. (2016). Assessing sensitivity to unmeasured confounding using a simulated potential confounder. *Journal of Research on Educational Effectiveness*, *9*(3), 395–420.

Chazan, D., Sela, H., & Herbst, P. (2016). Is the role of algebra in secondary school changing? *Journal of Curriculum Studies*, *48*(4), 506–528. doi: 10.1080/00220272.2015.1122091

Cinelli, C., & Hazlett, C. (2020). Making sense of sensitivity: Extending omitted variable bias. *Journal of the Royal Statistical Society: Series B (Statistical Methodology)*, *82*(1), 39–67.

Clotfelter, C. T., Ladd, H. F., & Vigdor, J. L. (2015). The aftermath of accelerating algebra: Evidence from a district policy initiative. *Journal of Human Resources*, *50*(1), 159–188. doi: 10.3368/jhr.50.1.159

Cohen, D. K., & Hill, H. C. (2000). Instructional policy and classroom performance: The mathematics reform in california. *Teachers college record*, *102*(2), 294–343.

Cui, Y., & Tchetgen Tchetgen, E. (2021). A semiparametric instrumental variable approach to optimal treatment regimes under endogeneity. *Journal of the American*

*Statistical Association*, *116*(533), 162–173.

Domina, T., McEachin, A., Penner, A., & Wimberly, L. (2015). Universal algebra and math achievement: Evidence from the district of columbia. *Educational Evaluation and Policy Analysis*, *37*(1), 70–92. doi: 10.3102/0162373714527783

Feodor Nielsen, S. (2000). The stochastic em algorithm: estimation and asymptotic results.

Hardt, M., Price, E., & Srebro, N. (2016). Equality of opportunity in supervised learning. *Advances in neural information processing systems*, *29*.

Jackson, J. W. (2021). Meaningful causal decompositions in health equity research: definition, identification, and estimation through a weighting framework. *Epidemiology*, *32*(2), 282–290.

Jackson, J. W., & VanderWeele, T. (2018). Decomposition analysis to identify intervention targets for reducing disparities. *Epidemiology*, *29*(6), 825–835.

Kalogrides, D., & Loeb, S. (2013). Different teachers, different peers: The magnitude of student sorting within schools. *Educational Researcher*, *42*(6), 304–316.

Kamiran, F., & Calders, T. (2012). Data preprocessing techniques for classification without discrimination. *Knowledge and information systems*, *33*(1), 1–33.

Kaufman, J. S. (2008). Epidemiologic analysis of racial/ethnic disparities: some fundamental issues and a cautionary example. *Social Science & Medicine*, *66*(8), 1659–1669.

Kelly, S. (2009). The black-white gap in mathematics course taking. *Sociology of Education*, *82*(1), 47–69.

Long, M. C., Conger, D., & Iatarola, P. (2012). Effects of high school course-taking on secondary and postsecondary success. *American Educational Research Journal*, *49*(2), 285–322.

Loveless, T. (2008). *The misplaced math student: Lost in eighth-grade algebra.* Washington, DC: Brookings Institution.

Lundberg, I. (2020). The gap-closing estimand: A causal approach to study interventions that close disparities across social categories. *Sociological Methods & Research*, 00491241211055769.

Murphy, S. A. (2003). Optimal dynamic treatment regimes. *Journal of the Royal Statistical Society: Series B (Statistical Methodology)*, *65*(2), 331–355.

National Center for Education Statistics. (2019). *High school longitudinal study of 2009 (hsls:09) first-year students' mathematics coursetaking and achievement* (Tech. Rep.). U.S. Department of Education. Retrieved from `https://nces.ed.gov/pubs2019/2019154/index.asp` (Retrieved from `https://nces.ed.gov/pubs2019/2019154/index.asp`)

Park, S., Kang, S., Lee, C., & Ma, S. (2023). Sensitivity analysis for causal decomposition analysis: Assessing robustness toward omitted variable bias. *Journal of Causal Inference*, *11*(1), 20220031.

Park, S., Lee, N., & Quintana, R. (2024a). Causal decomposition analysis with time-varying mediators: designing individualized interventions to reduce social disparities. *Sociological Methods & Research*, 00491241241264562.

Park, S., Lee, N., & Quintana, R. (2024b). Causal decomposition analysis with time-varying mediators: Designing individualized interventions to reduce social disparities. *Sociological Methods & Research*, *0*(0), 00491241241264562. doi: 10.1177/00491241241264562

Park, S., Qin, X., & Lee, C. (2022). Estimation and sensitivity analysis for causal decomposition in health disparity research. *Sociological Methods & Research*, 00491241211067516.

Pearl, J. (2009). *Causality: Models, reasoning, and inference* (Second ed.). Cambridge: Cambridge University Press.

Qian, M., & Murphy, S. A. (2011). Performance guarantees for individualized treatment rules. *Annals of statistics*, *39*(2), 1180.

Qin, X., & Yang, F. (2022). Simulation-based sensitivity analysis for causal mediation studies. *Psychological Methods*, *27*(6), 1000.

Qiu, H., Carone, M., Sadikova, E., Petukhova, M., Kessler, R. C., & Luedtke, A. (2021). Optimal individualized decision rules using instrumental variable methods. *Journal of the American Statistical Association*, *116*(533), 174–191.

Riegle-Crumb, C., & Grodsky, E. (2010). Racial-ethnic differences at the intersection of math course-taking and achievement. *Sociology of Education*, *83*(3), 248–270.

Robins, J. M., Hernan, M. A., & Brumback, B. (2000). Marginal structural models and causal inference in epidemiology. *Epidemiology*, 550–560.

Rosenbaum, J. E. (1999). If tracking is bad, is detracking better?. *American Educator*, *23*(4), 24.

Rubin, D. B. (1986). Statistical matching using file concatenation with adjusted weights and multiple imputations. *Journal of Business & Economic Statistics*, *4*(1), 87–94.

Rubin, D. B. (2004). *Multiple imputation for nonresponse in surveys* (Vol. 81). John Wiley & Sons.

Rubinstein, M., Branson, Z., & Kennedy, E. H. (2023). Heterogeneous interventional effects with multiple mediators: Semiparametric and nonparametric approaches. *Journal of Causal Inference*, *11*(1), 20220070.

Setoguchi, S., Schneeweiss, S., Brookhart, M. A., Glynn, R. J., & Cook, E. F. (2008). Evaluating uses of data mining techniques in propensity score estimation: a simulation study. *Pharmacoepidemiology and drug safety*, *17*(6), 546–555.

Shao, J. (1996). Bootstrap model selection. *Journal of the American statistical Association*, *91*(434), 655–665.

Silver, E. A. (1995). Rethinking algebra for all. *Educational Leadership*, *52*(6), 30–33.

Stein, M. K., Kaufman, J. H., Sherman, M., & Hillen, A. F. (2011). Algebra: A challenge at the crossroads of policy and practice. *Review of Educational Research*, *81*(4), 453–492.

Tsiatis, A. A., Davidian, M., Holloway, S. T., & Laber, E. B. (2019). *Dynamic treatment regimes: Statistical methods for precision medicine*. Chapman and Hall/CRC.

Ulrich, R., & Wirtz, M. (2004). On the correlation of a naturally and an artificially dichotomized variable. *British Journal of Mathematical and Statistical Psychology*, *57*(2), 235–251.

VanderWeele, T., & Robinson, W. R. (2014). On causal interpretation of race in regressions adjusting for confounding and mediating variables. *Epidemiology (Cambridge, Mass.)*, *25*(4), 473-483.

Zhang, B., Tsiatis, A. A., Davidian, M., Zhang, M., & Laber, E. (2012). Estimating optimal treatment regimes from a classification perspective. *Stat*, *1*(1), 103–114.

Zhang, Z. (2019). Reinforcement learning in clinical medicine: a method to optimize dynamic treatment regime over time. *Annals of translational medicine*, *7*(14).